\documentclass{article}
\usepackage{ijcai25}

\usepackage[marginal]{footmisc}
\usepackage{times}
\usepackage{soul}
\usepackage{url}
\usepackage[hidelinks]{hyperref}
\usepackage[utf8]{inputenc}
\usepackage[small]{caption}
\usepackage{graphicx}
\usepackage{amsmath}
\usepackage{amsthm}
\usepackage{booktabs}
\usepackage{algorithm}
\usepackage{algorithmic}
\usepackage[switch]{lineno}
\usepackage{subfigure}
\usepackage[edges]{forest}
\usetikzlibrary{arrows.meta,shapes,positioning,shadows,trees}
\usepackage{setspace}
\usepackage{color}
\usepackage{tikz}

\title{Logical Reasoning in Large Language Models: A Survey}

\author{
Hanmeng Liu$^1$\footnotemark[2]
\and
Zhizhang Fu$^1$\footnotemark[2] \and
Mengru Ding$^1$\and
Ruoxi Ning$^1$ \\
Chaoli Zhang$^2$ \and
Xiaozhang Liu$^3$ \And
Yue Zhang$^1$\footnotemark[1] \\
\affiliations
$^1$ Westlake University\\
$^2$ Zhejiang Normal University\\
$^3$ Hainan University\\ 
\emails
\{liuhanmeng, zhangyue\}@westlake.edu.cn,
\{fuzhizhang.fzz, dingmengru2021\}@gmail.com,
ruoxining@outlook.com,
chaolizcl@zjnu.edu.cn,
lxzh@hainanu.edu.cn
}

\begin{document}

\maketitle

\renewcommand{\thefootnote}{\fnsymbol{footnote}}
\footnotetext[1]{Corresponding author.}
\footnotetext[2]{Equal contribution.}

\begin{abstract}
    With the emergence of advanced reasoning models like OpenAI o3 and DeepSeek-R1, large language models (LLMs) have demonstrated remarkable reasoning capabilities. However, their ability to perform rigorous logical reasoning remains an open question. This survey synthesizes recent advancements in logical reasoning within LLMs, a critical area of AI research. It outlines the scope of logical reasoning in LLMs, its theoretical foundations, and the benchmarks used to evaluate reasoning proficiency. We analyze existing capabilities across different reasoning paradigms — deductive, inductive, abductive, and analogical — and assess strategies to enhance reasoning performance, including data-centric tuning, reinforcement learning, decoding strategies, and neuro-symbolic approaches. The review concludes with future directions, emphasizing the need for further exploration to strengthen logical reasoning in AI systems.
\end{abstract}

\section{Introduction}
\label{intro}

Logical reasoning is a fundamental challenge to artificial intelligence (AI) and natural language processing (NLP)~\cite{1056797,MCCARTHY1981431,McCarthy_Programs59}. While early formal logic-based reasoning approaches faced limitations in scalability and adaptability ~\cite{pereira1982logic,aa2411fb24e644d38091336c968e8de8}, data-driven models became the dominant method since the 1980s~\cite{mccarthy1989artificial}.
Recently, pre-trained Large Language Models (LLMs) and their emergent logical reasoning abilities have attracted increasing attention~\cite{liu2023evaluating,xu2023large}. Logical reasoning integrates LLMs with inference structuring, enabling multistep deduction and abstraction, and improving interpretability and reliability~\cite{shi-etal-2021-neural,stacey-etal-2022-logical,rajaraman-etal-2023-investigating}. It also strengthens generalization, helping models handle novel scenarios beyond their training data~\cite{haruta-etal-2020-logical}. As LLMs become integral to domains like legal analysis and scientific discovery, ensuring the correctness and verifiability of their reasoning is increasingly vital. As a result, post-training LLM for reasoning has garnered a surge of interest in both industry and research\cite{openai2024learning,deepseekai2024deepseekr1,muennighoff2025s1}. 

\definecolor{mycolor}{RGB}{215, 245, 200}
\tikzstyle{my-box}=[
    rectangle,
    draw=hidden-draw,
    rounded corners,
    text opacity=1,
    minimum height=1.5em,
    minimum width=5em,
    inner sep=2pt,
    align=center,
    fill opacity=.5,
]
\begin{figure*}[tp]
    \centering
    \resizebox{0.85\linewidth}{!}{\begin{forest}
            forked edges,
            for tree={
                grow=east,
                reversed=true,
                anchor=base west,
                parent anchor=east,
                child anchor=west,
                base=left,
                font=\small,
                rectangle,
                draw=violet,
                rounded corners,
                align=left,
                minimum width=4em,
                edge+={darkgray, line width=1pt},
                s sep=3pt,
                inner xsep=2pt,
                inner ysep=3pt,
                ver/.style={rotate=90, child anchor=north, parent anchor=south, anchor=center},
            },
            where level=1{
                text width=8em,
                font=\scriptsize,
            }{},
            where level=2{
                if n children=0
                    {text width=22.5em, font=\scriptsize, fill=mycolor}
                    {text width=11.2em, font=\scriptsize},
            }{},
            where level=3{
                if n children=0
                    {text width=34.4em, font=\scriptsize, fill=mycolor}
                    {text width=8em, font=\scriptsize},
            }{},
            where level=4{
                if n children=0
                    {text width=24.9em, font=\scriptsize, fill=mycolor}
                    {text width=10em, font=\scriptsize},
            }{},
            [
                LLM Logical Reasoning, ver
                [
                    Types \& history (\S \ref{sec:history})
                ]
                [
                    Task \& Benchmarks (\S \ref{sec:task })
                    [
                        Natural Language Inference (\S \ref{sec:nli })
                        [
                            ConTRoL \cite{Liu_Cui_Liu_Zhang_2021} {,} FOLIO~\cite{han-etal-2024-folio} {,}
                            LogicNLI \cite{tian-etal-2021-diagnosing} {,} \\ 
                            RulteTaker~\cite{10.5555/3491440.3491977} {,}
                            LogiBench \cite{parmar2023logicbench}
                        ]
                    ]
                    [
                        Reading Comprehension (\S \ref{sec:qa })
                        [
                            LogiQA \cite{10174688} {,} ReClor \cite{yu2020reclor} {,}
                            AR-LSAT \cite{yu2020reclor} {,} 
                            CLUTRR \cite{sinha2019clutrr} {,} \\
                            GSM \cite{cobbe2021gsm8k,li-etal-2024-gsm}
                            LINGOLY \cite{bean2024lingoly}
                        ]
                    ]
                    [
                        Benchmarks and test suites (\S \ref{sec:suite })
                        [
                            GLoRE \cite{liu2023glore} {,} LogiGLUE \cite{luo2024logiglue} {,}
                            LogiTorch \cite{helwe2022logitorch}
                        ]
                    ]
                ]
                [
                    Evaluation \& Analysis (\S \ref{sec:eval })
                    [
                        Deductive Reasoning (\S \ref{sec:deductive })
                        [
                            \cite{NEURIPS2023_09425891} {,} \cite{yuan-etal-2023-pretrained} {,} \cite{ryb-etal-2022-analog}
                        ]
                    ]
                    [
                        Inductive Reasoning (\S\ref{sec:inductive })
                        [
                            \cite{yang-etal-2024-language} {,} \cite{bowen-etal-2024-comprehensive} {,} \cite{sullivan-2024-true}
                        ]
                    ]
                    [
                        Abductive Reasoning (\S \ref{sec:abductive })
                        [
                            True Detective \cite{del-fishel-2023-true} {,} \cite{nguyen2023sota}
                        ]
                    ]
                    [
                        Analogical Reasoning (\S\ref{sec:analogical })
                        [
                            ANALOGICAL \cite{wijesiriwardene-etal-2023-analogical} {,}
                            \cite{petersen-van-der-plas-2023-language} {,} \cite{qin2024relevant}
                        ]
                    ]
                    [
                        Overall Analysis \& Metrics(\S\ref{sec:general })
                        [
                            \cite{liu2023evaluating} {,} \cite{xu2023large} {,} \cite{liu2024aligning} {,} \cite{gandarela2024inductive} {,} \cite{thatikonda2025assessing}
                        ]
                    ]
                ]
                [
                    Enhancement Methods (\S\ref{sec:method})
                    [
                        Data-Centric Approaches (\S\ref{sec:data-centric approaches})
                        [
                            Expert-Curated Datasets
                            [
                                FOLIO~\cite{han-etal-2024-folio} {,}
                                P-FOLIO~\cite{han-etal-2024-p} {,} \\
                                LeanDojo~\cite{10.5555/3666122.3667066} {,}
                                Symbol-LLM~\cite{xu-etal-2024-symbol}
                            ]
                        ]
                        [
                            Synthetic Datasets
                            [
                                RulteTaker~\cite{10.5555/3491440.3491977} {,}
                                \text{FLD$_{\times 2}$}~\cite{NEURIPS2024_8678da90}
                            ]
                        ]
                        [
                            LLM-distilled Datasets
                            [
                                LogiCoT~\cite{liu2023logicot} {,}
                                LogicPro~\cite{jiang2024logicpro} {,} \\
                                PODA~\cite{wang2024thought}
                            ]
                        ]
                    ]
                    [
                        Model-Centric Approaches   (\S\ref{sec:model-centric approaches})
                        [
                            Instruction Fine-Tuning
                            [
                                LogiCoT \cite{liu2023logicot} {,} LogiPT \cite{feng-etal-2024-language} {,} 
                                PGL \cite{wang2024efficient} {,} \\
                                Symbol-LLM \cite{xu-etal-2024-symbol} {,}
                                TPCL \cite{wang2024thought} {,}
                            ]
                        ]
                        [
                            Reinforcement Learning
                            [
                                \cite{jiao-etal-2024-learning} {,}
                                \cite{10.5555/3692070.3694287} {,}
                                Marco-o1~\cite{zhao2024marco} {,}
                                \\
                                Deepseek-R1-Zero~\cite{deepseekai2024deepseekr1} {,}
                                Deepseek-R1~\cite{deepseekai2024deepseekr1}
                            ]
                        ]
                        [
                            Inference-Time Decoding 
                            [
                                GoT~\cite{lei2023boosting} {,}
                                Chain of Logic~\cite{servantez-etal-2024-chain} {,} \\
                                Selection-Inference~\cite{DBLP:conf/iclr/CreswellSH23} {,}
                                \cite{malon2024exploring} {,}
                                \\
                                Maieutic Prompting~\cite{jung-etal-2022-maieutic} {,}
                                Logic-of-thought~\cite{liu2024logic} {,}
                                \\
                                DetermLR~\cite{sun2024determlr} {,}
                                Neurologic~\cite{lu-etal-2021-neurologic} {,}
                                \\
                                Formal-LLM~\cite{li2024formal}
                            ]
                        ]
                    ]  
                    [
                        External Knowledge Utilization (\S\ref{sec:external knowledge utilization})
                        [
                            \cite{zayyad2024formallanguageknowledgecorpus} {,}
                            LeanDojo~\cite{10.5555/3666122.3667066} {,}
                            LQOT~\cite{liu2024logicquerythoughtsguiding} {,}
                            \cite{ouyang2023factdrivenlogicalreasoningmachine} {,} \\
                            KnowRA~\cite{mai2025knowraknowledgeretrievalaugmented}
                        ]
                    ]
                    [
                        Neuro-Symbolic Approaches
                        (\S\ref{sec:neuro-symbolic})
                        [
                            LINC~\cite{olausson-etal-2023-linc} {,}
                            LOGICLLAMA~\cite{yang-etal-2024-harnessing} {,} \\
                            CLOVER~\cite{ryu2024divide} {,}
                            LOGIC-LM~\cite{pan-etal-2023-logic} {,}
                            Logic Agent~\cite{liu2024logic} {,} \\
                            LLM-TRes~\cite{toroghi2024verifiable} {,}
                            SymbCoT~\cite{xu-etal-2024-faithful} {,}
                            Aristotle~\cite{xu2024aristotle}
                        ]
                    ]
                ]
                [
                     Discussion (\S\ref{sec:discussion})
                ]
            ]
        \end{forest}}
    \caption{The structure of this survey}
    \label{fig:taxo2}
    \vspace{-1.5em}
\end{figure*}
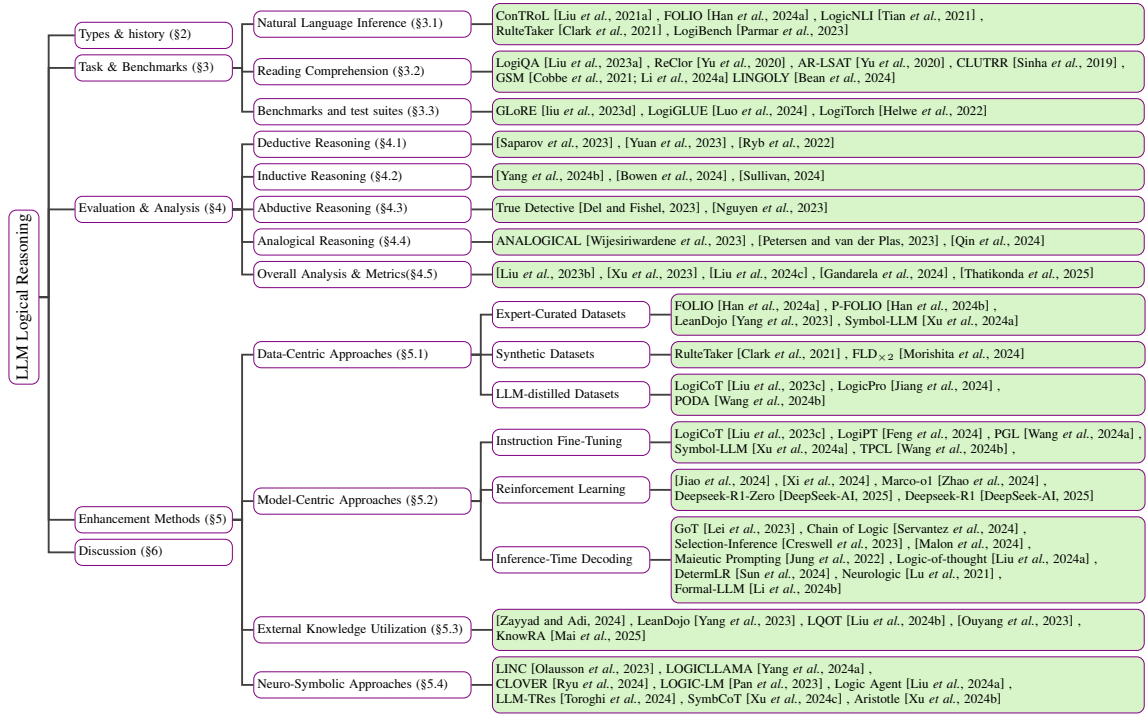

Despite growing research, existing surveys~\cite{plaat2024reasoning,sun2023survey,yu2024natural} often conflate logical reasoning with general-purpose heuristic strategies like Chain-of-Thought (CoT)~\cite{xia2024beyond}. There has been a lack of a literature review dedicated to LLM and formal symbolic logic. This survey provides a comprehensive review of logical reasoning in large language models (LLMs), with a focus on formal and symbolic logic-based reasoning rather than general heuristic approaches. We begin by defining logical reasoning in AI, distinguishing it from general-purpose reasoning, and categorizing key paradigms, including deductive, inductive, abductive, and analogical reasoning. Additionally, we analyze existing benchmarks and evaluation methodologies, identifying gaps in assessing symbolic inference, consistency, and robustness. We further explore state-of-the-art techniques for enhancing logical reasoning, such as instruction fine-tuning, logic-informed pre-training, reinforcement learning, inference-time decoding strategies, and hybrid neuro-symbolic methods. 
We examine recent advances in neuro-symbolic integration, along with applications of theorem provers, logic solvers, and formal verification frameworks in LLMs. Finally, we highlight open challenges in scalability, reasoning consistency, explainability, and efficiency, proposing future directions for multi-modal reasoning, hybrid architectures, and improved evaluation frameworks. The structure of the subsequent chapters is illustrated in Figure~\ref{fig:taxo2}.

\section{Logic in Artificial Intelligence} 
\label{sec:history}

Logical reasoning is a cornerstone of artificial intelligence (AI), enabling machines to simulate human thought processes and solve complex problems. At its core, logical reasoning applies structured rules to derive conclusions from premises, providing a rigorous framework for decision-making and inference~\cite{sun2023survey}.

\subsection{History of Logic Reasoning Research}  
\label{sec:development}
Logical reasoning can be traced back to ancient Greece, where Aristotle’s syllogisms laid the foundation for classical logic. During the Middle Ages, scholars refined these theories, and in the 17th century, Leibniz’s universal language and calculus ratiocinator bridged logic with mathematics, foreshadowing modern computational logic. The 19th century saw George Boole’s Boolean algebra, which transformed logic into a mathematical framework, laying the foundation for digital computing.

The 20th century ushered in modern logic, with Russell and Whitehead’s Principia Mathematica formalizing complex logical systems. By the mid-century, AI pioneers like John McCarthy leveraged logic for knowledge representation and automated theorem proving, leading to logic programming and knowledge bases. The 1970s introduced non-monotonic logic, enabling AI to handle commonsense reasoning. The 1980s saw logical reasoning integrate with knowledge representation, advancing expert systems for real-world applications. The 1990s saw the rise of knowledge graphs, structuring vast knowledge for complex reasoning tasks.

In the 21st century, neuro-symbolic approaches have merged deep learning with logical inference, resulting in tools like DeepLogic~\cite{cingillioglu2019deeplogic} and SATNet~\cite{wang2019satnet}. Logical reasoning remains a cornerstone of AI research, evolving from philosophy to modern computing. As AI advances, logical reasoning continues to shape intelligent systems, ensuring structured, interpretable, and robust decision-making.

\subsection{Types of Logical Reasoning}  
\label{sec:definition}

Logical reasoning can be broadly categorized into four main types, each serving distinct purposes and applications:

\paragraph{Deductive Reasoning.} This type of reasoning derives specific conclusions from general principles or premises. It operates under the rule that if all premises are true and the reasoning is valid, the conclusion must also be true. For example, given the premises ``All apples are red" and ``This fruit is an apple" one can deduce that ``This fruit is red" Deductive reasoning is fundamental in fields such as mathematics and formal logic, where certainty and rigor are paramount.

\paragraph{Inductive Reasoning.} Unlike deductive reasoning, inductive reasoning draws general conclusions based on specific observations or evidence. While the conclusions are often considered probable, they are not guaranteed to be true. For instance, observing that all swans seen so far are white might lead to the inductive conclusion that ``All swans are white" Inductive reasoning is widely used in scientific discovery and data-driven decision-making, where patterns and trends are inferred from empirical data.

\paragraph{Abductive Reasoning.} This form of reasoning seeks the most plausible explanation or cause for a set of observations, often in the presence of incomplete information. Abductive reasoning is particularly useful in diagnostic tasks and real-world problem-solving. For example, seeing wet spots on the street might lead one to infer that ``It has recently rained" While abductive conclusions are not certain, they provide a practical basis for hypothesis generation and decision-making under uncertainty.

\begin{table}[htbp]
\centering
\resizebox{\linewidth}{!}{\begin{tabular}{l|c|c|c|c}
\hline
\textbf{Dataset} & \textbf{Language} & \textbf{Question Type} & \textbf{Size} & \textbf{Source} \\
\hline
LogiQA & Zh/En & Multichoice & 15,937 & Exam-based \\
ReClor & En & Multichoice & 6,138  & Exam-based \\
AR-LSAT & En & Multichoice & 2,064 & Exam-based\\
CLUTRR & En & Question-answering & 6,016 & Rule-based\\
GSM & En & Math word problems & 19K
& Exam-based \\
LINGOLY & En & Question-answering & 1,133 & Expert-designed\\
\hline
ConTRoL & En & ternary classification & 8,325 & Exam-based\\
FOLIO & En & binary classification & 1,351 & Expert-designed \\
LogicNLI & En & ternary classification & 30K & Exam-based \\
ProofWriter & En & binary classification & - & Exam-based \\
LogicBench & En & binary classification & 1,270 & Rule-based \\
\hline
GLoRE & Zh/En & Miscellaneous & 17 tasks & Miscellaneous \\
LogiGLUE & En & Miscellaneous & 24 tasks & Miscellaneous \\
LogiTorch & En & Miscellaneous & 16 tasks & Miscellaneous \\
BIG-Bench & En & Miscellaneous & 7 tasks & Miscellaneous \\

\hline
\end{tabular}}
\caption{Main Datasets and Benchmarks of Logical Reasoning Task.}
\label{tab:datasets}
\vspace{-1.5em}
\end{table}

\paragraph{Analogical Reasoning.} Analogical reasoning involves drawing comparisons between similar situations or domains to make inferences or solve problems. By identifying parallels between different scenarios, this type of reasoning enables creative problem-solving and knowledge transfer. For example, understanding that planets orbit the sun in elliptical paths might lead one to analogically reason that other celestial bodies, such as comets, exhibit similar orbital characteristics. Analogical reasoning is particularly valuable in fields like education, design, and innovation.

\section{Tasks and Benchmarks}
\label{sec:task }

Logical reasoning datasets and benchmarks are essential for evaluating the reasoning capabilities of large language models (LLMs). These datasets can be categorized into three types based on their data sources:

\textbf{Rule-based Datasets}~\cite{tafjord-etal-2021-proofwriter,sinha2019clutrr} are automatically generated using logical rules, enabling large-scale data collection. However, ensuring diversity is crucial to avoid repetitive patterns and comprehensively evaluate reasoning capabilities.

\textbf{Expert-Designed Datasets}~\cite{han-etal-2024-folio} are constructed by domain experts, ensuring high precision and accuracy. Although typically smaller than crowd-sourced corpora, their meticulous design makes them indispensable for in-depth logical reasoning evaluation.

\textbf{Exam-Based Datasets}~\cite{liu2020logiqa,yu2020reclor,wang2022lsat} originate from standardized test questions (e.g., Chinese National Civil Service Exam, LSAT, GRE), offering high-quality, expert-crafted logic problems at scale. These datasets are widely used to evaluate reasoning in real-world scenarios.

Table~\ref{tab:datasets} summarizes important datasets for logical reasoning, which typically cover tasks such as Natural Language Inference (NLI) (\S\ref{sec:nli }) and Machine Reading Comprehension (MRC) (\S\ref{sec:qa }).

\subsection{Natural Language Inference (NLI)}
\label{sec:nli }

NLI evaluates whether a \textit{hypothesis} logically follows from a \textit{premise}, directly assessing a model’s reasoning ability. Labels typically fall into binary (Entailment, Non-entailment) or ternary (Entailment, Contradiction, Neutral) classifications. Some datasets use \emph{True} and \emph{False} labels instead.

\textbf{ConTRoL}~\cite{Liu_Cui_Liu_Zhang_2021} is derived from recruitment exams (e.g., bank entry, U.S. police selection), containing 8,325 entries with \emph{Correct}, \emph{Incorrect}, and \emph{Can't Say} labels, corresponding to \emph{Entailment}, \emph{Contradiction}, and \emph{Neutral}.

\textbf{FOLIO}~\cite{han-etal-2024-folio} is an expert-constructed dataset for First-Order Logic (FOL) reasoning, consisting of 1,351 entries labeled as \emph{True} or \emph{False}, making it a rigorous benchmark for formal logical inference.

\textbf{LogicNLI}~\cite{tian-etal-2021-diagnosing} contains 30K entries generated using logical rules, with \emph{Entailment}, \emph{Contradiction}, and \emph{Neutral} labels. It isolates FOL-based inference from commonsense reasoning, enabling precise evaluation of reasoning accuracy and generalization.

\textbf{ProofWriter}~\cite{tafjord-etal-2021-proofwriter} extends RuleTaker~\cite{10.5555/3491440.3491977} by introducing CWA (closed-world assumption) and OWA (open-world assumption) to handle negation and open-world reasoning. It includes Birds-Electricity (handcrafted domain theories) and ParaRules (crowdsourced paraphrased rules) for systematic evaluation of generalization across linguistic variations and real-world knowledge domains.

\textbf{LogicBench}~\cite{parmar2023logicbench} is a GPT-3-generated dataset covering 25 types of reasoning, including propositional logic, FOL, and non-monotonic logic. It consists of 1,270 test entries labeled as \emph{Yes} or \emph{No}.

\subsection{Machine Reading Comprehension (MRC)}
\label{sec:qa }

MRC evaluates logical reasoning by requiring models to answer questions based on a given passage. Tasks are commonly formatted as multiple-choice, span extraction, or free response, with multiple-choice QA being particularly effective due to its standardization.

\textbf{LogiQA}~\cite{10174688} is sourced from the Chinese Civil Service Exam, containing 15,937 entries in Chinese and English. It targets complex logical reasoning and is widely used for evaluating LLMs.

\textbf{ReClor}~\cite{yu2020reclor}, derived from the GMAT, features 6,138 English entries with four-option multiple-choice questions.

\textbf{AR-LSAT}~\cite{wang2022lsat} is based on the LSAT, containing 2,064 entries spanning ordering games, grouping games, and allocation games, each with five options.

\textbf{CLUTRR}~\cite{sinha2019clutrr} focuses on inductive reasoning, requiring models to infer kinship relationships in short narratives. It contains 6,016 entries, combining entity extraction and logical inference.

\textbf{GSM} evaluates mathematical reasoning capabilities, comprising two datasets: GSM8K~\cite{cobbe2021gsm8k} (8.5K grade school math problems) and GSM-PLUS~\cite{li-etal-2024-gsm} (10,552), which is augmented with mathematical perturbations for robustness evaluation.

\textbf{LINGOLY}~\cite{bean2024lingoly} uses Linguistic Olympiad puzzles to evaluate in-context pattern identification and generalization in low-resource or extinct languages. It contains 1,133 problems across 6 formats and 5 difficulty levels, covering over 90 languages.

\subsection{Benchmark Suites}
\label{sec:suite }

Benchmark suites standardize evaluation and facilitate model comparison in logical reasoning research.

\textbf{GLoRE}~\cite{liu2023glore} is a few-shot and zero-shot testing platform, including 17 test-only datasets to assess generalization in low-data scenarios.

\textbf{LogiGLUE}~\cite{luo2024logiglue} consists of 24 logical reasoning tasks, standardizing datasets into a sequence-to-sequence format for uniform input processing. It provides both test and training sets, enabling extensive model training and targeted evaluations.

\textbf{LogiTorch}~\cite{helwe2022logitorch} is a PyTorch-based library for natural language logical reasoning, offering 16 datasets, model architectures, and an accessible API for quick evaluation.

\textbf{BIG-bench}~\cite{srivastava2022beyond} is a collaborative benchmark with 7 tasks dedicated to logical reasoning, such as Logic Grid Puzzle and Logical Fallacy Detection.

\begin{figure}[htbp]
    \centering
    \subfigure[A multi-choice reading comprehension example from the LogiQA dataset.]{
        \centering
        \label{fig:logiqa}
        \includegraphics[width=0.45\textwidth]{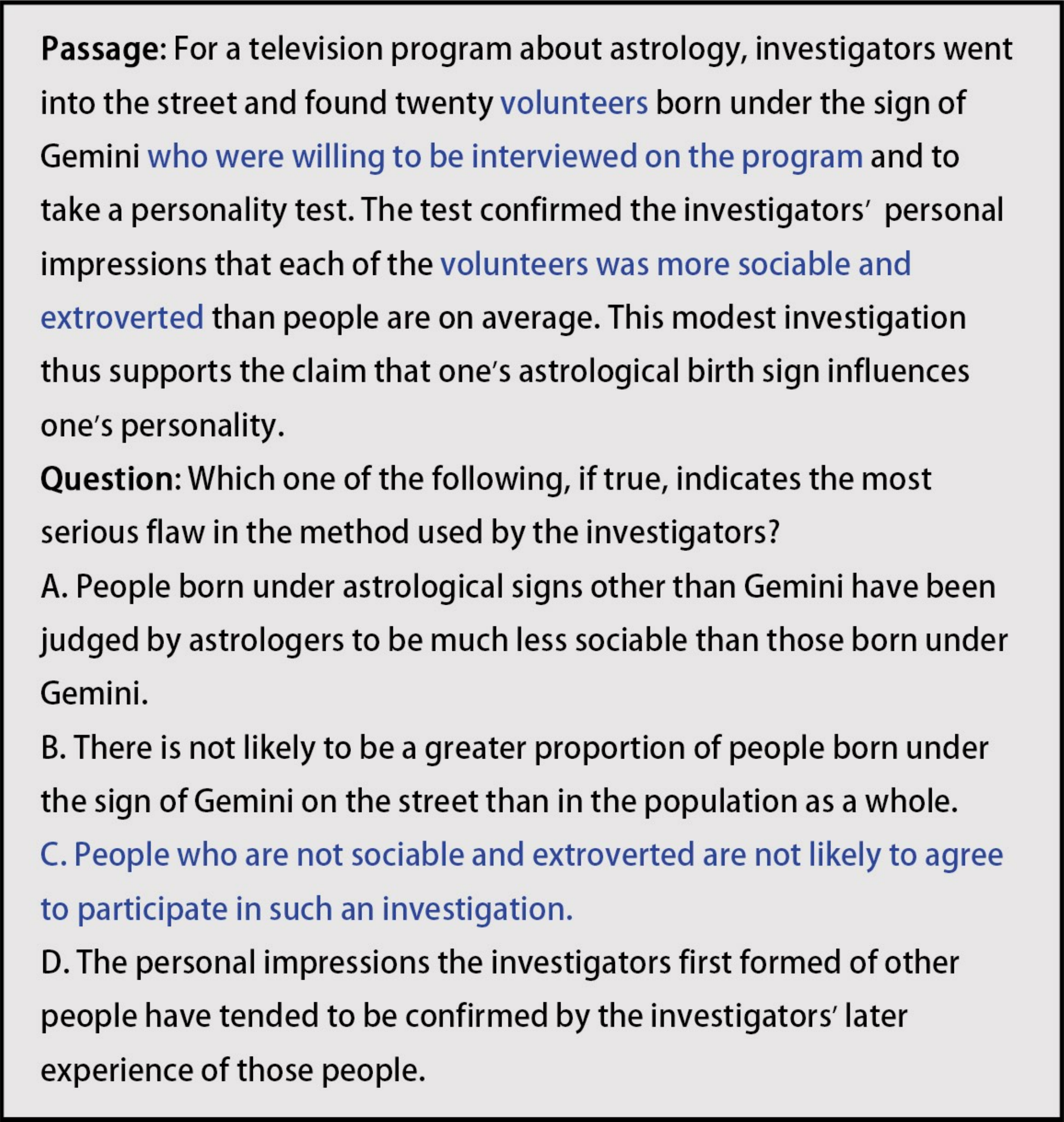} 
        }
        \vspace{-0.1em}
    \subfigure[An NLI example from the ConTRoL dataset.]{
        \centering
        \label{fig:control}
        \includegraphics[width=0.45\textwidth]{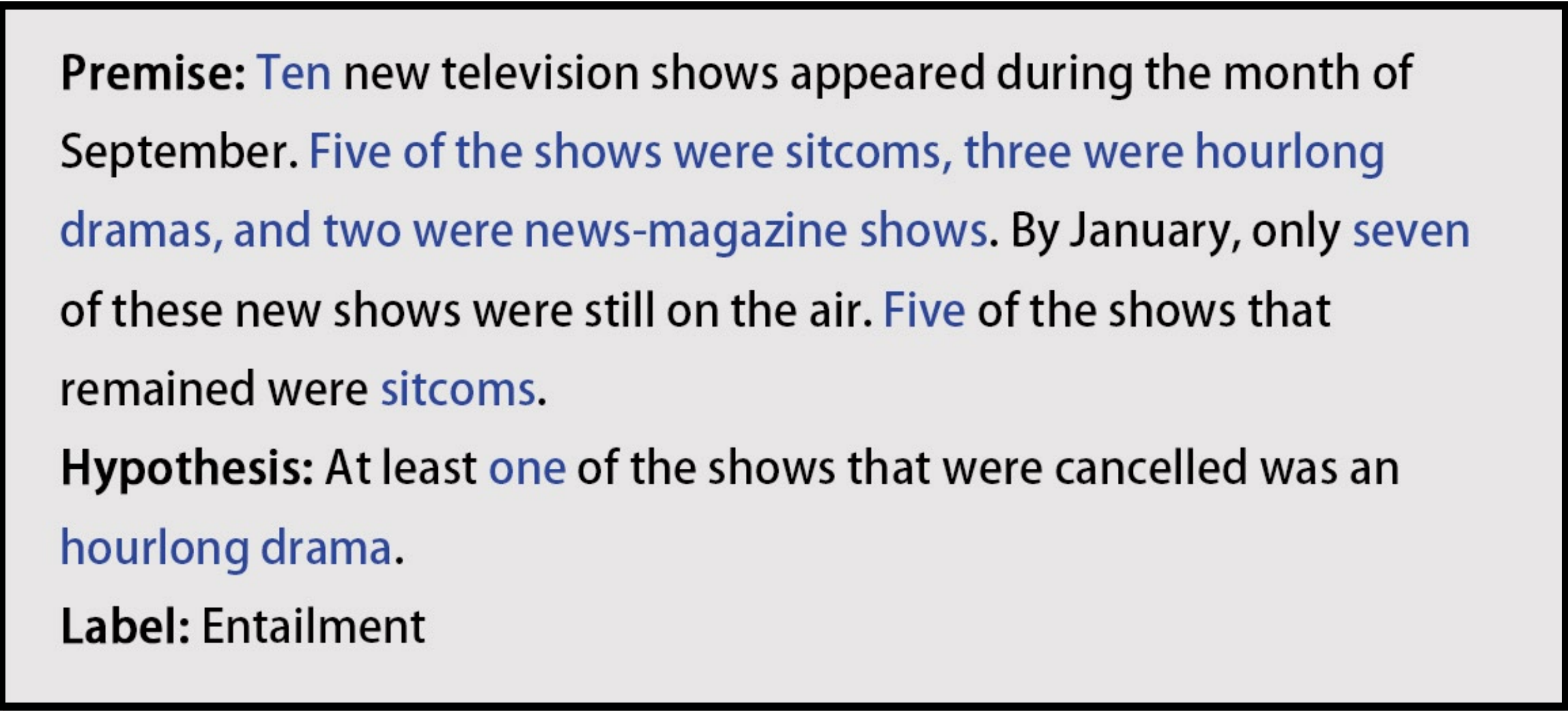}
    }
    \vspace{-0.5em}
    \caption{Example tests of Logical reasoning in NLP tasks.} 
    \label{fig:examples}
    \vspace{-2em}
\end{figure}

\section{Evaluations}
\label{sec:eval }
The rapid development of pre-trained language models (PLMs) necessitates rigorous evaluation of their logical reasoning capabilities. This section examines four reasoning paradigms—deductive, inductive, abductive, and analogical—while analyzing evaluation approaches and metrics.

\subsection{Deductive Reasoning}
\label{sec:deductive }

Deductive reasoning, deriving specific conclusions from general premises, is crucial for automated theorem proving. Despite LLMs performing well on tasks like compositional proofs, standard benchmarks, and encoding entailment relationships, they struggle with extended reasoning, hypothetical sub-proofs without examples, generalization, and sensitivity to syntactic variations~\cite{NEURIPS2023_09425891,yuan-etal-2023-pretrained,ryb-etal-2022-analog}.



\subsection{Inductive Reasoning}
\label{sec:inductive }

Inductive reasoning, which generalizes from specific instances to broader rules, is essential for tasks like hypothesis generation and pattern recognition. While \citeauthor{yang-etal-2024-language}~\shortcite{yang-etal-2024-language} find that pre-trained models can serve as effective ``reasoners,'' \citeauthor{bowen-etal-2024-comprehensive}~\shortcite{bowen-etal-2024-comprehensive} show that even advanced LLMs struggle with simple inductive tasks in their symbolic settings. Similarly, \citeauthor{sullivan-2024-true}~\shortcite{sullivan-2024-true} demonstrates that Transformer models, even after fine-tuning, fail to learn fundamental logical principles, indicating limited inductive reasoning capabilities.



\subsection{Abductive Reasoning}
\label{sec:abductive }

Abductive reasoning, which seeks the most plausible explanations for observed phenomena, is crucial in fields like law and medicine. \citeauthor{del-fishel-2023-true}~\shortcite{del-fishel-2023-true} highlights the challenges LLMs face in generating plausible hypotheses from incomplete information. In the legal domain, \citeauthor{nguyen2023sota}~\shortcite{nguyen2023sota} show that despite strong performance, models struggle with abductive reasoning, underscoring the complexity of this paradigm.


\subsection{Analogical Reasoning}
\label{sec:analogical }

Analogical reasoning, which infers unknown information by comparing it with known information, is vital for tasks requiring creativity and knowledge transfer. \citeauthor{wijesiriwardene-etal-2023-analogical}~\shortcite{wijesiriwardene-etal-2023-analogical} introduced ANALOGICAL, a benchmark for long-text analogical reasoning. They find that as analogy complexity increases, LLMs struggle to recognize analogical pairs. \citeauthor{petersen-van-der-plas-2023-language}~\shortcite{petersen-van-der-plas-2023-language} show that models can learn analogical reasoning with minimal data, approaching human performance. However, \citeauthor{qin2024relevant}~\shortcite{qin2024relevant} question whether LLMs truly rely on analogical reasoning, discovering that random examples in prompts often achieve comparable performance to relevant examples.

\subsection{Overall Analysis and Metrics}
\label{sec:general }

\citeauthor{liu2023evaluating}~\shortcite{liu2023evaluating} evaluate GPT-4 and ChatGPT on benchmarks like LogiQA and ReClor, showing that while GPT-4 outperforms ChatGPT, both of them struggle with out-of-distribution tasks. \citeauthor{xu2023large}~\shortcite{xu2023large} introduce the NeuLR dataset and propose a framework evaluating LLMs across six dimensions: correctness, rigor, self-awareness, proactivity, guidance, and absence of hallucinations.

\paragraph{Metrics for Evaluating Logical Reasoning.}
\label{sec:metrics}

Traditional metrics like accuracy and F1 score are insufficient for assessing logical reasoning. Recent studies have introduced nuanced metrics such as consistency (invariance to logically equivalent inputs), generalization (performance on out-of-distribution data), and explainability (clarity of reasoning steps). \citeauthor{thatikonda2025assessing}~\shortcite{thatikonda2025assessing} find that combining BERTScore with traditional metrics improves alignment with human judgments. \citeauthor{liu2024aligning}~\shortcite{liu2024aligning} propose a framework for measuring logical consistency, showing that BERTScore aligns better with human rankings than LLM-based evaluators like GPT-4. \citeauthor{gandarela2024inductive}~\shortcite{gandarela2024inductive} emphasizes the need for metrics that account for the expressivity of logical theories, particularly in inductive reasoning.

\section{Enhancement Methods} 
\label{sec:method}


Enhancing LLMs' logical reasoning remains crucial. This section focuses on core strategies: Data-Centric Approaches (\S\ref{sec:data-centric approaches}), Model-Centric Approaches (\S\ref{sec:model-centric approaches}), External Knowledge Utilization (\S\ref{sec:external knowledge utilization}), and Neuro-Symbolic Reasoning (\S\ref{sec:neuro-symbolic}).

\subsection{Data-Centric Approaches} 
\label{sec:data-centric approaches}

Data-centric approaches enhance LLMs' reasoning capabilities by utilizing meticulously curated training datasets. Formally, this can be expressed as:  

\begin{equation}  
D^* = \arg\max_{D} R(M_D) 
\end{equation}  

where:  
\begin{itemize}  
    \item \( D \): training datasets.
    \item \( M_D \): model trained on \( D \).
    \item \( R \): performance evaluator (e.g., LLM-as-a-judge, rule-based metrics).
\end{itemize}  

This formulation highlights the central role of dataset optimization in data-centric approaches. In practice, data-centric methods typically involve three types of datasets: expert-curated datasets, synthetic datasets, and LLM-distilled datasets.




\paragraph{Expert-Curated Datasets.} The FOLIO series~\cite{han-etal-2024-folio,han-etal-2024-p} establish formal verification through FOL annotations, with P-FOLIO extending the complexity of reasoning chains for enhanced training. LeanDojo~\cite{10.5555/3666122.3667066} provides 98k+ human-proven theorem pairs for mathematical reasoning. Additionally, Symbol-LLM~\cite{xu-etal-2024-symbol} systematically organizes 34 symbolic reasoning tasks to capture inter-symbol relationships across 20 distinct symbolic families.

\paragraph{Synthetic Datasets.} Rule-based synthetic data remains fundamental for data generation. RuleTaker~\cite{10.5555/3491440.3491977} formalizes this through a three-phase pipeline: behavior formalization, example synthesis and linguistic equivalents generation. Similarly, \citeauthor{NEURIPS2024_8678da90}~\shortcite{NEURIPS2024_8678da90} develops Formal Logic Deduction Diverse (\text{FLD$_{\times 2}$}), a synthetic dataset based on symbolic theory and previous empirical insights.

\paragraph{LLM-Distilled Datasets.} Researchers employ advanced models such as GPT-4 for intermediate reasoning step distillation. LogiCoT~\cite{liu2023logicot} augments existing datasets with GPT4-generated reasoning chains, while LogicPro~\cite{jiang2024logicpro} combines algorithmic problems with code solutions to create variable-guided reasoning data. To advance, \citeauthor{wang2024thought}~\shortcite{wang2024thought} propose PODA, which generates contrastive analyses of correct/incorrect options through premise-oriented augmentation, enabling reasoning path differentiation via contrastive learning.

\subsection{Model-Centric Approaches} 
\label{sec:model-centric approaches}
Model-Centric approaches enhance LLMs' reasoning capabilities by optimizing model parameters and decoding strategies. The formal objective is:  

\begin{equation}  
(\theta^*, S^*) = \arg\max_{\theta, S} R(M_{\theta}, S)  
\end{equation}  

where:  
\begin{itemize}  
    \item $\theta$: learnable model parameters.
    \item $M_\theta$: model with parameters $\theta$.
    \item $S$: decoding strategy (e.g., chain-of-thought prompting, verification-based decoding).
    \item $R$: reasoning performance metric. 
\end{itemize}  

This formulation highlights the joint optimization of model parameters \( \theta \) and decoding strategy \( S \). Practical implementations can be categorized as:  
\begin{itemize}  
    \item Instruction Fine-Tuning: optimizing \( \theta \).  
    \item Reinforcement Learning: optimizing \( \theta \).  
    \item Inference-Time Decoding: optimizing \( S \).   
\end{itemize}

Model-Centric approaches focus on directly improving the model's reasoning capabilities by optimizing its internal mechanisms and decoding strategies, making them complementary to data-centric approaches.

\subsubsection{Instruction Fine-Tuning}
Instruction Fine-Tuning (IFT) adapts LLMs through supervised learning on task-specific instructions. For example, \citeauthor{liu2023logicot}~\shortcite{liu2023logicot} design multi-grained instructions spanning diverse levels of abstraction and complexity. Similarly, \citeauthor{feng-etal-2024-language}~\shortcite{feng-etal-2024-language} IFT models to mimic logical solvers by replicating formal deduction reasoning processes. In addition, \citeauthor{xu-etal-2024-symbol}~\shortcite{xu-etal-2024-symbol} implement two-stage symbolic fine-tuning through \textit{Injection} (injecting symbolic knowledge) and \textit{Infusion} (balancing symbol and NL reasoning).

To overcome IFT's over-fitting limitations, 
\citeauthor{wang2024thought}~\shortcite{wang2024thought} enforce contrastive learning between factual/counterfactual paths with IFT. Further, \citeauthor{wang2024efficient}~\shortcite{wang2024efficient} augment Llamas with a Program-Guided Learning Framework and logic-specific architecture adjustments.

Recently, \citeauthor{muennighoff2025s1}~\shortcite{muennighoff2025s1} propose s1, achieving test-time scaling through IFT on 1,000 meticulously crafted long CoT samples. Combined with budget-forcing technique, it significantly enhances the reasoning capability of a Qwen2.5-32B-Instruct model, allowing extrapolating beyond its performance without test-time intervention.

\subsubsection{Reinforcement Learning}
Reinforcement learning (RL) has become pivotal in optimizing large language models (LLMs), particularly since the breakthrough of Reinforcement Learning from Human Feedback (RLHF). \citeauthor{jiao-etal-2024-learning}~\shortcite{jiao-etal-2024-learning} leverage RL for planning-based reasoning optimization, while \citeauthor{10.5555/3692070.3694287}~\shortcite{10.5555/3692070.3694287} develop 
$R^3$, achieving process supervision benefits through outcome-only supervision.

The success of large-scale RL in OpenAI-o1~\cite{openai2024learning} has inspired numerous studies. RL algorithms train o1-style models to enhance Chain-of-Thought (CoT) reasoning, addressing issues like formulaic outputs and limited long-form reasoning. For instance, \citeauthor{zhao2024marco}~\shortcite{zhao2024marco} integrate CoT instruction fine-tuning with Monte Carlo Tree Search (MCTS) decoding for multi-path reasoning exploration. In contrast, \citeauthor{zhang2024o1}~\shortcite{zhang2024o1} employ MCTS to generate code-reasoning data for instruction fine-tuning (IFT) and Direct Preference Optimization (DPO).

A significant breakthrough comes from DeepSeek-R1~\cite{deepseekai2024deepseekr1}, which pioneers a novel RL strategy to enhance logical reasoning. DeepSeek-R1-Zero, trained purely through RL without IFT, demonstrates impressive reasoning capabilities but faces challenges in readability and language consistency. To address this, DeepSeek-R1 introduces minimal long-CoT IFT data as a cold start before RL, achieving a balance between usability and reasoning performance. By iteratively synthesizing high-quality reasoning data through RL, DeepSeek-R1 overcomes limitations imposed by human annotators, addressing issues such as mechanistic responses, repetitive patterns, and insufficient long-chain reasoning. This approach represents a potential paradigm shift in logical reasoning optimization, pushing the boundaries of what LLMs can achieve in structured reasoning tasks.

\subsubsection{Inference-Time Decoding}
 We categorize logical reasoning enhancement methods during inference-time into inference-time scaling and constrained decoding.

Inference-time scaling employs computational augmentation without parameter updates. One common approach is decoding with structured outputs and modular workflows. GoT~\cite{lei2023boosting} creates structured reasoning nodes to improve complex multi-step logical reasoning. Similarly, Chain of Logic~\cite{servantez-etal-2024-chain} introduces a Decomposition-Recomposition structure for legal reasoning. In other contexts, researchers design more complex modular workflows for better performance~\cite{DBLP:conf/iclr/CreswellSH23,malon2024exploring}.

Another inference-time scaling approach involves stimulating autonomous reasoning, guiding LLMs to iteratively refine their answers. Maieutic Prompting~\cite{jung-etal-2022-maieutic} eliminates contradictions through recursive reasoning. Similarly, Logic-of-Thoughts~\cite{liu2024logic} and DetermLR~\cite{sun2024determlr} progressively approach the answers in an iterative style.

Constrained decoding methods, on the other hand, focus on improving the controllability and reliability of reasoning processes. Neurologic~\cite{lu-etal-2021-neurologic} enforces predicate logic constraints, while Formal-LLM~\cite{li2024formal} integrates automata for constraining plan generation.

\subsection{External Knowledge Utilization} 
\label{sec:external knowledge utilization}
LLMs often generate incorrect answers due to hallucinations when performing complex tasks such as logical reasoning, making it necessary to incorporate external knowledge to assist in producing accurate responses. Formally, the optimal integration of external knowledge can be formulated as a joint optimization problem:

\begin{equation}  
(M^*, K^*) = \arg\max_{M, K} R(M, K)  
\end{equation}  

where:  
\begin{itemize}  
    \item \( M \): the neural model, which includes both the model's parameters and its decoding strategies (generally, the model's parameters remain unchanged).
    \item \( K \): knowledge integration strategy, including knowledge source curation, structured knowledge representation, retrieval-augmented mechanisms, etc.
    \item \( R \): reasoning performance evaluator (e.g., factual accuracy, logical consistency).
\end{itemize} 

\citeauthor{zayyad2024formallanguageknowledgecorpus}~\shortcite{zayyad2024formallanguageknowledgecorpus} and \citeauthor{10.5555/3666122.3667066}~\shortcite{10.5555/3666122.3667066} extract data from Lean, a mathematical proof tool, to aid theorem proving. In contrast, ``Logic-Query-of-Thoughts" (LQOT)~\cite{liu2024logicquerythoughtsguiding} decomposes complex logical problems into easier sub-questions before integrating knowledge graphs.

In reading comprehension, \citeauthor{ouyang2023factdrivenlogicalreasoningmachine}~\shortcite{ouyang2023factdrivenlogicalreasoningmachine} construct supergraphs to address complex contextual reasoning, while KnowRA~\cite{mai2025knowraknowledgeretrievalaugmented} autonomously determines whether to accept external knowledge to assist document-level relation extraction.

\subsection{Neuro-Symbolic Approaches} 
\label{sec:neuro-symbolic}
Neural-symbolic hybrid methods represent a burgeoning research area that aims to combine the powerful representational capabilities of deep learning with the precision and interpretability of symbolic reasoning. 

Formally, a neural-symbolic hybrid system aims to optimize both the neural model \( M \) and the symbolic solver \( P \) (where \( P \) represents the symbolic reasoning process) to maximize logical reasoning performance. The overall objective can be expressed as:  

\[  
(M^*, P^*) = \arg\max_{M, P} R(P(M(x))),  
\]  

where:  
\begin{itemize}  
    \item \( M \): The neural model, which includes both the model's parameters and its decoding strategies. It maps the input \( x \) (e.g., natural language) into a symbolic representation \( z \) within a formal language \( \mathcal{L} \):  
    \[  
    z = M(x), \quad z \in \mathcal{L}.  
    \]  
    \item \( P \): The symbolic solver, which operates on the symbolic representation \( z \) produced by \( M \) to generate the final output \( y \):  
    \[  
    y = P(z).  
    \]  
    \item \( R \): The reasoning performance metric, which evaluates the ability to perform logical reasoning tasks. 
\end{itemize}  

The optimization process involves two key directions:  
\begin{itemize}  
    \item Improving \( M \): including refining the model's parameters and decoding strategies to produce symbolic representations that are both accurate and compatible with \( P \).
    \item Enhancing \( P \): involving improving the symbolic solver's ability to process.
\end{itemize}  

By jointly optimizing \( M \) and \( P \), neural-symbolic hybrid systems aim to leverage the strengths of both neural networks and symbolic reasoning to achieve superior logical reasoning capabilities. It is worth noting that in earlier neural-symbolic pipelines, \( P \) is often implemented as a fixed external logical reasoning engine, and thus is generally not optimized. However, in advanced practice, LLMs are increasingly being used to perform the role of \( P \), enabling diverse optimization.

Fundamentally, these methods involve translating problems into symbolic representations with LLMs, and external symbolic solvers solving them. For example, in LINC~\cite{olausson-etal-2023-linc}, LLMs convert natural language (NL) into first-order logic (FOL) expressions, and utilize an external theorem prover for symbolic deductive inference.

Further efforts focus on improving NL-to-symbolic translation. 
One prevailing approach is directly optimizing translation through training~\cite{yang-etal-2024-harnessing} or decoding strategies~\cite{ryu2024divide}, while the other depends on verification or correction mechanisms~\cite{yang-etal-2024-harnessing,pan-etal-2023-logic}.

Building upon these, recent advancements address the traditional pipeline limitations by fully integrating LLMs into reasoning processes. Logic Agent (LA)~\cite{liu2024logic} replaces external solvers with rule-guided LLM inference chains, while LLM-TRes~\cite{toroghi2024verifiable} implements self-contained verifiable reasoning without external symbolic solvers. SymbCoT~\cite{xu-etal-2024-faithful} coordinates translation, planning, solving and verification entirely through LLMs. \citeauthor{xu2024aristotle}~\shortcite{xu2024aristotle} propose Aristotle, which further systematizes the symbolic reasoning pipeline through three LLM-driven components: Logical Decomposer, Logical Search Router, and Logical Resolver.

\section{Discussion}
\label{sec:discussion}

The integration of logical reasoning into large language models (LLMs) remains a critical challenge, marked by persistent gaps between heuristic performance and formal logical rigor. Below, we analyze three unresolved tensions dominating the field and outline future directions.

\paragraph{Robustness vs. Generalization.}
LLMs exhibit inconsistent performance in structured reasoning tasks such as deductive inference and abductive hypothesis generation. While models fine-tuned on datasets like FOLIO~\cite{han-etal-2024-folio} excel in controlled settings, they struggle with adversarial perturbations or semantically equivalent rephrasings. This inconsistency arises from their reliance on surface-level statistical correlations rather than causal relationships, coupled with limited out-of-distribution generalization. A key question persists: can LLMs achieve human-like robustness without sacrificing cross-domain adaptability? Current methods prioritize narrow task performance, leaving real-world applicability uncertain.

\paragraph{Interpretability vs. Performance.}
A central tension lies in balancing neural scalability with symbolic precision. Neuro-symbolic approaches like Logic-LM~\cite{pan-etal-2023-logic} and Symbol-LLM~\cite{xu-etal-2024-symbol} embed formal logic solvers into neural architectures, improving interpretability through step-by-step proofs. However, these methods face scalability bottlenecks with large knowledge bases or complex rule dependencies. Conversely, data-driven methods (e.g., instruction tuning on LogicBench~\cite{parmar2024logicbench}) achieve broader task coverage but fail to generalize beyond syntactic patterns. How can we reconcile transparent reasoning with black-box model performance? Hybrid architectures offer promise but introduce computational overhead, limiting practical deployment.

\paragraph{Evaluation Rigor.}
Existing benchmarks like LogiQA~\cite{liu2020logiqa} and ReClor~\cite{yu2020reclor} conflate reasoning ability with pattern recognition through multiple-choice formats. While efforts like NeuLR~\cite{xu2023large} curate ``neutral" content to isolate reasoning from domain knowledge, they lack scope for holistic evaluation. Current metrics (e.g., accuracy, BLEU) fail to assess consistency (invariance to logically equivalent inputs) or soundness (adherence to formal proof structures). What defines a gold standard for logical reasoning evaluation? Benchmarks must prioritize systematic testing of core principles (e.g., transitivity, contraposition) over task-specific performance.

\paragraph{Future Directions.}
Addressing these challenges requires hybrid architectures that dynamically integrate neural and symbolic components, such as differentiable theorem provers, to balance scalability and precision. Equally important is the development of evaluation frameworks that stress-test models on perturbed logical statements (e.g., negated premises, swapped quantifiers) to isolate reasoning from memorization. Multimodal reasoning, which grounds inference in diverse modalities (text, images, code), presents untapped potential for enhancing robustness and interpretability. Finally, interdisciplinary collaboration—leveraging insights from formal logic, cognitive science, and machine learning—will be essential to design systems that reason with and about uncertainty. Until LLMs reliably disentangle logic from lexicon, their deployment in high-stakes domains will remain precarious. Bridging this gap demands rigorous benchmarks, scalable hybrid methods, and a redefinition of evaluation paradigms.

\section{Conclusion}
This survey synthesizes the rapid advancements and persistent challenges in logical reasoning for large language models (LLMs). While LLMs demonstrate impressive heuristic reasoning, rigorous logical inference—spanning deductive, inductive, abductive, and analogical paradigms—remains inconsistent due to limitations in robustness, generalization, and interpretability. We analyzed strategies to enhance reasoning, including neuro-symbolic integration, data-centric tuning, reinforcement learning, test-time scaling, and other improved decoding methods, and highlighted benchmarks like FOLIO and LogiQA for systematic evaluation. Future progress hinges on hybrid architectures that unify neural and symbolic reasoning, robust evaluation frameworks, and scalable methods for cross-domain and multimodal inference. Addressing these challenges will advance LLMs toward reliable, interpretable reasoning critical for real-world applications.

\clearpage
\begin{spacing}{0.3}

    \fontsize{7.8}{11}\selectfont
    \bibliographystyle{named}
    \bibliography{ijcai25}
\end{spacing}

\end{document}